\documentclass[10pt, a4paper, onecolumn]{article} 
\usepackage[margin=0.75in]{geometry} 

\usepackage{cite}
\usepackage{amsmath,amssymb,amsfonts}
\usepackage{algorithmic}
\usepackage{graphicx}
\usepackage{textcomp}
\usepackage{xcolor}

\usepackage[font=footnotesize]{subcaption}
\usepackage[font=footnotesize]{caption}

\usepackage{xurl} 

\usepackage{hyperref} 
\hypersetup{pdfborder={0 0 0}} 
\usepackage{orcidlink} 

\usepackage{authblk} 

\begin{document}

\title{
(hu)Man \textit{vs.} Machine: In the Future of Motorsport, can Autonomous Vehicles Compete?
}

\author[1]{Armand Amaritei \orcidlink{0009-0000-2810-2314}}
\author[2]{Amber-Lily Blackadder \orcidlink{0009-0003-5153-0132}}
\author[1]{Sebastian Donnelly \orcidlink{0009-0009-9716-3503}}
\author[1]{Lora Hernandez \orcidlink{0009-0003-5153-0132}}
\author[2]{James Vine \orcidlink{0009-0009-2546-805X}}
\author[3]{Alexander Rast \orcidlink{0000-0001-9934-7191}}
\author[2]{Matthias Rolf \orcidlink{0000-0003-0563-3264}}
\author[1]{Andrew Bradley \orcidlink{0000-0001-7053-804X}}

\affil[1]{Autonomous Driving and Intelligent Transport Group, Oxford Brookes University, UK}
\affil[2]{School of Engineering, Computing \& Mathematics, Oxford Brookes University, UK}
\affil[3]{Artificial Intelligence, Data Analysis and Systems Institute, Oxford Brookes University, UK}

\affil[ ]{\textit{\{19205454, arast, mrolf, abradley\}@brookes.ac.uk}}

\date{}

\maketitle

\begin{abstract}
Motorsport has historically driven technological innovation in the automotive industry. Autonomous racing provides a proving ground to push the limits of performance of autonomous vehicle (AV) systems. In principle, AVs could be at least as fast, if not faster, than humans. However, human driven racing provides broader audience appeal thus far, and is more strategically challenging. Both provide opportunities to push each other even further technologically, yet competitions remain separate. This paper evaluates whether the future of motorsport could encompass joint competition between humans and AVs. Analysis of the current state of the art, as well as recent competition outcomes, shows that while technical performance has reached comparable levels, there are substantial challenges in racecraft, strategy and safety that need to be overcome. Outstanding issues involved in mixed human-AI racing, ranging from an initial assessment of critical factors such as system-level latencies, to effective planning and risk guarantees are explored. The crucial non-technical aspect of audience engagement and appeal regarding the changing character of motorsport is addressed. In the wider context of motorsport and AVs, this work outlines a proposed agenda for future research to 'keep pushing the possible', in the true spirit of motorsport.
\end{abstract}

\smallskip
\section{Introduction}
\smallskip
Autonomous vehicles (AVs) have been rising in popularity \cite{alvarezleonIndustryEmergenceMarket2022} driving the success of many companies, \textit{e.g}. Tesla and Waymo. Current studies \cite{waymo_2023_report} show that Waymo AVs are much safer than human drivers, reducing the risk of a serious injury by upto 90\%. Meanwhile, the concept of racing robots against each other is not new, with competitions such as Micromouse and Roboracer. As AVs keep rising in popularity, another question can now be asked, when will it be possible for autonomous race cars to race against human drivers? 

In this paper we seek to evaluate the current state of autonomous racing and explore the question: are mixed autonomous and human driven races the future of motorsport? The concept of ‘(hu)Man vs Machine’, and machines being 'better' than humans has a long history. Similarities can be drawn to other competitions such as chess, where it is widely accepted that computers are better than human players, being capable of beating top-rated human grandmasters since 1997 when the reigning world champion Gary Kasparov was defeated for the first time by a computer \cite{modiEffectsComputerAI2023}. Far from signalling the 'death' of chess as a competitive game, this has presented human players with the opportunity to learn from how machines play to improve their own decision making. The future of motorsport could follow a similar trajectory, as autonomous racing cars have the potential to be faster and much more consistent than professional human drivers, since they have access to much more information of their surroundings, while not having to consider factors such as G-forces, fatigue, or, to some extent, risk. This essentially would create a ‘perfect’ driver which professional human drivers could learn from to optimise their lap times around a circuit. 

In the following sections, we outline the current state of AVs  in motorsport and attempt to predict how the future of motor sport could look like given recent data and technological advancements.
\smallskip
\section{Related Work}
\smallskip
\subsection{Autonomous Racing Competitions}
\smallskip
The emergence and growth of autonomous driving has been paralleled by a similar trajectory in autonomous racing. As autonomous competitions have gained more traction in recent years, there are now numerous events held around the world which contain some form of AVs racing against each other. 
\smallskip
\subsubsection{Abu Dhabi Autonomous Racing League}
\smallskip
The competition which has gained the most attention recently is the Abu Dhabi Autonomous Racing League (A2RL), which is the only racing event which races multiple AVs wheel-to-wheel around a circuit. The vehicles used in the event are modified Super Formula cars which allow for the mounting of an AI system and sensors. The sensor suite on the vehicles is standardised \cite{hoffmannHeadtoHeadAutonomousRacing2026a}, and stands out by its comprehensiveness compared to other common autonomous racing setups, containing 3 LiDARs, 4 RADAR sensors and 7 cameras. To allow for improved localisation, a GNSS receiver with an integrated Inertial Measurement Unit (IMU) and a motion sensor for visual velocity measurements are also included. The vehicles reach over 200 km/h, making high power compute capacity a critical requirement. This is provided by a 16 core AMD CPU and an NVIDIA RTX 6000 Ada GPU to run the autonomous stack.

While also competing against other AVs on track, the A2RL also has a segment which compares times from a human driver to the autonomous teams. In the first event which was held in 2024, the autonomous teams were, at their best, 10 seconds slower than professional driver, Daniil Kvyat \cite{A2RL2025}. This first season of the event also lacked the wheel-to-wheel racing element that is expected from motor sport. There were instances of AVs stopping on track when coming in close proximity with another vehicle and other AVs queuing up behind them, not being able to overtake.

The second season of this competition saw a drastic improvement in both the lap times of AVs and also the racing element. The Technical University of Munich (TUM) were only 1.5 slower \cite{BornToEngineer2025} on their fastest lap compared to Daniil Kvyat. There have also been significant improvements in relation to the overtaking aspect of the autonomous systems, with teams performing high-risk overtakes over the course of the race \cite{Raceteq2025}.
\smallskip
\subsubsection{Formula Student Germany}
\smallskip
There are similar competitions held in Europe, with Formula Student Germany (FSG) being one of the most recognised competitions. In this competition, AVs are not racing against each other on a track, but rather in a time-based competition along simpler tracks than in the A2RL, denoted by blue and yellow cones \cite{FSG_2026_Rules}.

FSG is historically a human-driven event, however, the driverless class was introduced in 2016 where teams had the chance to gain points by designing vehicles that were capable of switching between autonomous and manual driving modes. This meant teams had to balance high performance mechanical systems, with complex new software, whilst still ensuring reliability. Although not as fast as A2RL, these cars are able to reach speeds of over 100km/h across a 75 metre distance in the acceleration event \cite{FSG_About}. 
\smallskip
\subsubsection{Formula Student UK AI}
\smallskip
Another recognised competition in Europe is Formula Student UK AI (FSUK-AI). Similarly to A2RL, a standardized vehicle is provided to contenders, but sensors and compute equipment are mounted by each team separately. These custom ‘sensor plates’ fit onto the standardised low voltage vehicles which are shared by teams. These vehicles have a top speed of around 30 km/h \cite{FSAI_DRAFT_SPEC}. 
The rules of the competition and the specific race formats in which teams compete in are very similar to FSG \cite{IMechE2026}. 
This competition can be seen more as an entry point for teams who are planning to develop autonomous systems due to the lower speeds and more structured environments that it offers.
\smallskip
\subsection{Autonomous Racing Software Architecture}
\smallskip
There are similarities between all the architectures of the sensor suites and autonomous systems used in all three of the mentioned competitions. In terms of sensors, teams use similar types of sensors, typically comprising camera, LiDAR and an INS to be able to infer the position of the vehicle and identify the boundaries of the track.

From a software architecture perspective, teams in each competition generally follow a standard robotics pipeline, consisting of perception, cognition and actuation \cite{culleySystemDesignDriverless2020a}. A comparison was conducted between the systems of a team from each competition, Oxford Brookes Racing Autonomous (OBRA) \cite{culleySystemDesignDriverless2020a} from FSUK-AI, Akademischer Motorsportverein Zürich (AMZ) \cite{kabzanAMZDriverlessFull2020} from FSG and TUM \cite{hoffmannHeadtoHeadAutonomousRacing2026a} from A2RL. The systems of OBRA and AMZ are structured to achieve similar goals: their perception systems are built to detect blue, yellow, and orange cones which their Simultaneous Localisation and Mapping (SLAM) algorithms create a map from. SLAM results feed into path planning and then into an actuation node, which calculates what actuation commands the car required in order to follow the line given by the path planner. TUM’s system contains the same overall modules, albeit on a higher level of complexity, as they not only operate on a conventional (i.e. cone-less) circuit at higher speeds, but they also need to be aware and keep track of competitor vehicles. To make this possible, additional sensors are used - eliminating blind spots around the vehicle. This means that they are processing much more information than a human driver if placed in the same chassis. Their system moves away from the more ‘real-time’ aspect of the OBRA and AMZ systems - since they have a full map of the track before the race begins (similar to how an experienced driver may ‘learn the track’ before competing).

Notably, all these systems involve a function pipeline, which in computational terms introduces delays in the sensor-to-actuator system which are critical for performance. Although there are several works detailing the architecture of such software stacks \cite{culleySystemDesignDriverless2020a} \cite{kabzanAMZDriverlessFull2020} \cite{hoffmannHeadtoHeadAutonomousRacing2026a} there has been less work done to examine the latencies. A critical consideration for human-AV racing is an analysis of these latencies in the context of race results.
\smallskip
\section{Lessons from Formula SAE}
\smallskip
\subsection{Skidpad Performance, Human vs AI}
\smallskip
As an open series of events intended to develop students towards careers in engineering and motorsport, the Formula Student (FS) competition has generated a wealth of publicly-available data and insights on best practice. The FS competition involves (separate) automated and human driven timed events which are aimed at establishing the performance of student built vehicles and custom automated driving systems. 

\begin{figure} [ht]
    \centering
    \includegraphics[width=0.75\linewidth]{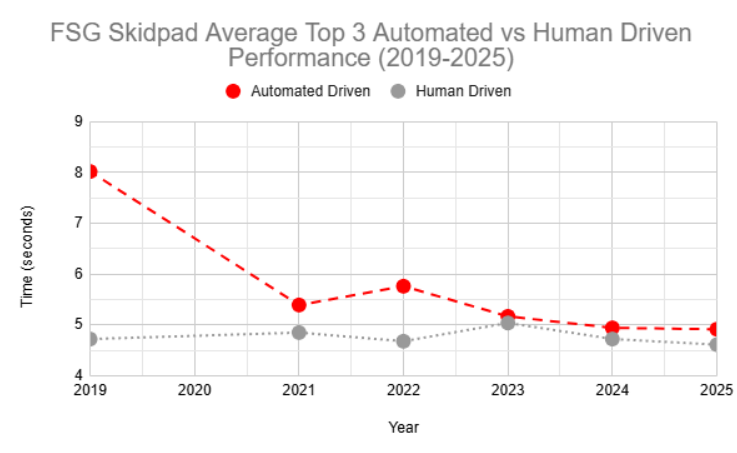}
    \caption{Comparison of human and automated driven performance over the past decade based on data for the Skidpad event at Formula Student Germany}
    \label{fig:skidpad_times}
\end{figure}

Of the four driven events at FS, skidpad is particularly suitable for the comparison of the performance of automated and human driven vehicles as FS Germany has been operating an automated vehicle category since 2017 \cite{FSG_About}, representing nearly a decade of development. The skidpad is a well-defined track setup of two circles with a radius of 10.75m which are driven in opposing directions to evaluate how the vehicle performs under high lateral acceleration - originally designed to test the capability of the vehicle itself at the limits of grip, however this challenge serves well to ascertain how the autonomous control system is able to physically maintain control of the vehicle at high lateral accelerations. At FSG 2019 the average skidpad circle completion time of the top 3 automated vehicles on the skidpad event was 70\% slower compared with the average for the top 3 human driven vehicles at the same event. By 2025 this reduced to just 6.6\%, representing a dramatic increase in automated vehicle performance in just over half a decade, while the human driven performance remained mostly unchanged (Fig.~\ref{fig:skidpad_times}).

\smallskip
\subsection{System Latency vs Human Latency} \label{latency}
\smallskip
Whilst far from being the only factor in motorsport performance \cite{doubekWhatMakesGood2021}, reaction times, or in the case of an automated driving system, end-to-end latency, have a significant impact on the ability to maintain control of a vehicle at the high speeds involved in racing \cite{betzAnalysisSoftwareLatency2023}. End-to-end latency affects the time it takes (and therefore distance travelled) to respond to an unexpected situation - in FS, this could be \textit{e.g.} suddenly encountering a knocked-over cone - on the open racetrack this could be rapidly responding to a competitor’s driving behaviour.

The end-to-end latency of OBRA’s software stack for the 2025 FSAI competition was measured to be 277ms. Whilst a direct comparison with human performance is difficult to assess due to the disparities in planned versus reactive sensing and action selection in the brain (\textit{e.g.}  in emergency braking, drivers might be able to react significantly faster by utilising previous track experience \cite{Jahfari2012}), research shows typical human reaction times while driving a car are in the range of 600-700ms \cite{culikEvaluationDriversReaction2022a}. Looking at simplified reaction times involving a button press response to visual stimuli, elite racing drivers demonstrate significantly lower reaction times in the order of 330ms compared with the 370ms of other participants \cite{baurReactivityStabilityStrength2006b}, and some anecdotal evidence suggests it may go as low as 200 ms \cite{f1_rapid}. The current OBRA system therefore clearly exceeds the tested reaction times of racing drivers. The comparison from FSG (Fig.~\ref{fig:skidpad_times}), indeed, supports this - if one may expect that the FSG human drivers are trained in racing but may not be at elite level, the data from 2023 onwards would suggest a similar closing of the gap in response latency. 
\smallskip
\subsection{Path-planning along a race track}
\smallskip
The path autonomous race cars take directly dictates the potential of how quick they can traverse the circuit. However, teams need to make a decision if they would rather take a safe centre line instead of riskier racing lines. 

In FSUK-AI, many teams choose to take a centre line around a circuit for simplicity and to decrease the chance of knocking over cones which would result in a time penalty or completely leaving the track, which would invalidate their run. 

\begin{figure} [hb]
    \centering
    \includegraphics[width=0.75\linewidth]{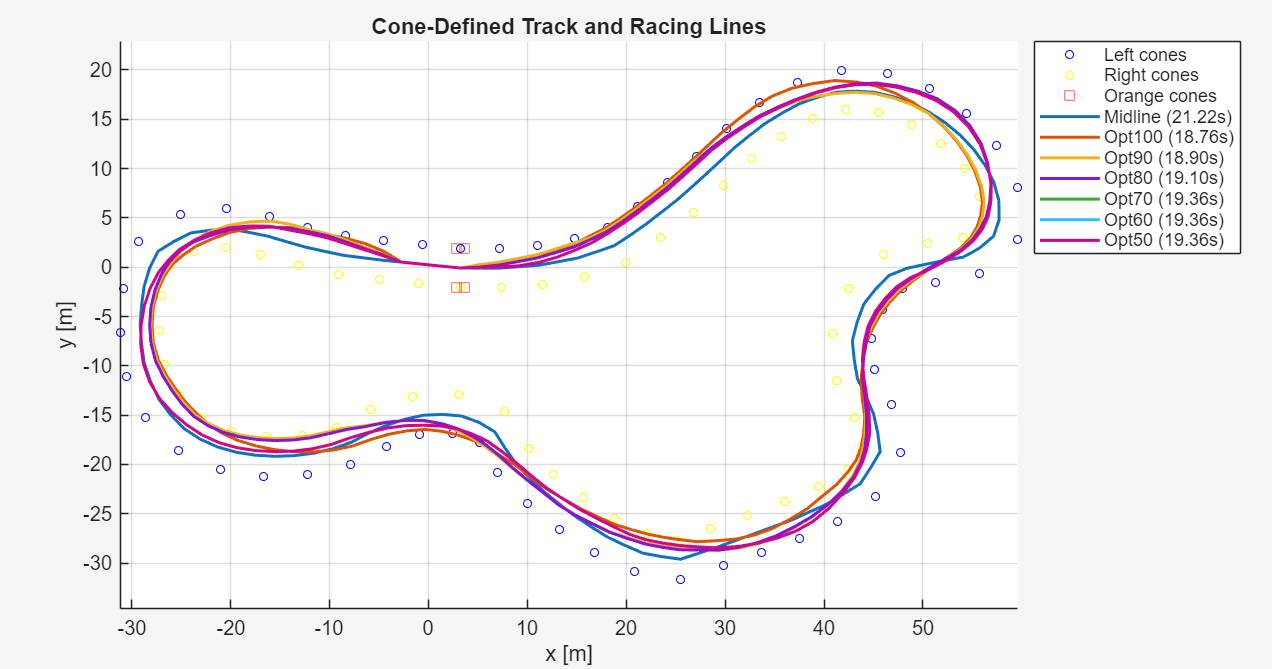}
    \caption{Comparison of timed laps along the mid-line and possible racing lines of a track with the ADS-DV (FSUK-AI).}
    \label{fig:racing_lines}
\end{figure}

With recent advances in racing line generation methods, the use of this in competition is now being explored for real time application in autonomous racing \cite{garlickRealtimeOptimalTrajectory2022}, \cite{heilmeierMinimumCurvatureTrajectory2020} to close the gap between human and AI driven vehicles. In this study, it was found that following the racing line (rather than the central path) has the potential to reduce the lap time by approximately 9.8\% (two seconds for the circuit shown in Fig.~\ref{fig:racing_lines}). The circuits used in the FS events are approximately 3 metres wide - thus there is potential for the lap time reduction to be even greater on a traditional race track where the full track width can be utilised to maximise the benefit of taking the racing line.

\newpage
\section{Remaining Obstacles \& Future Directions}
\smallskip
\subsection{Safety Considerations}
\smallskip
In the current formats, AV-vs-human racing has shown that AV technology has advanced towards human-level performance on an open race track. It has been able to perform on par with humans in time trials \cite{alcalaHumanMachineGap2024}, achieved in controlled environments such as staggered race starts, blue-flag passing protocols, conservative safety envelopes and enforced speed caps. However, such constrained environments and rules are not truly representative of a head-to-head race.‘Mixed racing’ entails direct interaction between drivers, and thus far, prevailing safety regulations have not, in general, articulated a coherent rule set for such conditions.

Wheel-to-wheel racing introduces human-AV interaction, tactical positioning, aggressive manoeuvres and intent uncertainty. Multi-agent interactions increase complexity in mixed racing, with the (autonomous) driver having to understand and predict competitor manoeuvres - not only to avoid collisions, but also to position themselves tactically for an optimal racing line, given the position of other vehicles \cite{zotero-item-643}. Furthermore, there are cases where the line itself may be chosen adversarially, to thwart the opponent (\textit{e.g.} to avoid being overtaken by a faster driver etc.), rather than simply optimising the ego-vehicle trajectory. To achieve this, autonomous systems must move away from minimising lap time under largely deterministic conditions toward optimising under dynamic and adversarial multi-agent environments. Optimal behaviour will no longer be defined by time efficiency, but by context-sensitive decision-making that maximises performance within the bounds of safety regulations.

Such an optimisation process depends strongly on the intrinsic capabilities of the autonomous system. Limitations including end-to-end latency, sensor resolution and state estimation accuracy restrict how quickly and reliably the AV can interpret and react to its surroundings (Section \ref{latency}). Computational latencies restrict the prediction horizon, model complexity, and trajectory planning that can be executed in real time. Control actuation delays constrain how precisely planned trajectories and manoeuvres can be executed. In addition, latencies limit behaviour and intent prediction of competitor drivers. Under high-speed conditions, where vehicles are pushed to the edge of their handling capabilities, small miscalculations and delayed outputs increase the risk and cost of collisions.

Safety-performance trade-offs emerge as performance optimisation in mixed racing occurs under uncertainty and strict safety constraints. Aggressive strategies that maximise lap-time efficiency and positional gain typically require reduction of safety margins, whereas conservative strategies preserve safety constraint at the cost of competitiveness. To maintain safety, AVs typically increase following distance, avoid marginal overtakes, reduce speed and enforce conservative collision-risk bounds. For example, in A2RL racing, the AV reacts to an approaching human-driven car by maintaining a safe distance. This is a tactical opportunity for human drivers to preemptively occupy a specific line around a corner or when overtaking. 

While current AVs are typically designed using probabilistic components in both the perception and actuation pipelines, they generally do not explicitly model risk. While new developments in AV perception have started to explore explicit modelling of uncertainty, this remains a research area \cite{manchingalEpistemicDeepLearning2025}. Experienced racing drivers routinely accept probabilistic risk, exploit competitor errors and dynamically adjust behaviour at the limits of tyre adhesion \cite{zotero-item-643} based on a learned internal model of uncertainty. Such decision-making integrates knowledge from subtle cues like vehicle feel (haptic feedback), prior knowledge of opponent behaviour, and contextual race dynamics that may not be easily articulated. AVs typically rely upon sensor inputs and predictive models to estimate vehicle state, environment condition and opponent trajectories. These models may not fully capture and predict the cues that humans interpret in real time - particularly where processing latencies are prevalent.
\smallskip
\subsection{Future Directions}
\smallskip
Equivalating human reaction times with end-to-end computational latency might not be adequate due to the fundamental difference between the way humans and machines process information \cite{kortelingHumanArtificialIntelligence2021}. Most studies on human reaction times focus on the reaction to simple visual or auditory stimuli rather than complex actions, something machines can react to in microseconds. When considering the instantaneous reaction to information presented with no prior information humans require up to 6-8s to take control of a vehicle \cite{zeebWhatDeterminesTakeover2015} while a machine can provide a complex, no-context control output within at most a few hundred milliseconds. The exact quantification of the reaction time of an experienced racing driver who has a clear idea of what could happen next on track, effectively correcting a mental model should be the focus for future research.

Some progress in risk-aware trajectory planning has been made in the normal road driving domain \cite{taghavifarOptimalReinforcementLearning2024}, examining risk as a predictive process - rather than a static estimation of instantaneous risk. However, the objective in this case is risk minimisation as opposed to risk optimisation - thus to adapt to the autonomous racing case, the model would have to predict not only risk but also potential reward. Such an approach naturally suggests reinforcement-learning approaches. 

Reinforcement learning is famously used to simulate cooperative and adversarial multi-agent interactions \cite{wernerDynamicMultiTeamRacing2023}, \cite{liuFormulaEMultiCarRace2024}. Together with game theoretic \cite{kalariaARACERRealTimeAlgorithm2025},\cite{linScenariobasedDecisionmakingUsing2025}, \cite{jiaRAPIDAutonomousMultiAgent2023} and Machine Theory of Mind \cite{chandraStylePredictMachineTheory2020} frameworks, they address the challenge of negotiating immediate interactions against competitors with different goals and intentions while considering long-term race-level objectives. These approaches would allow AVs to assess the value of taking an immediate action against the value of receiving a future potential reward – \textit{"do I risk an aggressive overtake now, risk the possibility of a collision, or keep myself in the race and aim for a high finishing position?"} These enable risk-based decision-making in AV’s.
\smallskip
\subsection{Maintaining the 'Character of Motorsport'}
\smallskip
A core aspect of motorsport is its ability to attract wide and diverse audiences through its elements of speed, risk, elite skill and competition. Racing audiences invest emotionally not only in the team but the individual driver, following narratives shaped by personality, resilience and risk taking - leading to the recent popularity of Netflix’s Drive to Survive. Therefore the character of motorsport is not only defined by speed but endurance, resilience, manually and mechanically\cite{MotorsportEntertainmentIndustry}. Fundamentally being a man-and-machine discipline, the character of motorsport is unlike any other sport, having the team extend beyond the drivers but to the wider technical team and machinery. The “team” includes designers, engineers monitoring telemetry in real time, strategists running simulations, and specialists optimising mechanical and software systems. As a result, the success is already shared between human and technological contributors, with drivers typically using language like \textit{“we won the race”} to preempt questions such as \textit{“whose glory is it - the driver, engineers or the car?”}. With AVs added to the mix, the question becomes at once more ambiguous and more intriguing - fodder for future debate and commentary.

The concept of "Man vs Machine" has long piqued human interest - leading to numerous fictional books and movies \textit{e.g.} RoboCop, Terminator, Blade Runner, and motivating researchers to build game-playing algorithms to challenge (and eventually defeat) humans in other competitive scenarios (\textit{e.g.} chess \cite{mullerManVsMachine2018}). Extending this concept to the complex, and hotly competitive, scenario of the racetrack provides an opportunity to excite and motivate the next generation of motorsport enthusiasts - potentially gaining support from the robotics community.

On the other hand, reducing the role of the human driver may diminish the perceived level of risk, weakening the emotional investment fans make to the sport, teams and drivers \cite{fullerStakeholderPerceptionsRisk2001}. Historically, the central appeal of motorsport has been driver skill, split-second decision making and willingness to push the limits - if safety concerns regarding autonomous systems result in a conservative, risk-averse rule set, then much of this appeal is diminished. Conversely, a human driver’s perception of risk is not solely based upon risk of losing (a position / race\textit{ etc}.) - there is also a consideration of the risk of personal injury. Since an autonomous driver cannot ‘hurt itself’, it has the potential to execute riskier manoeuvres than its human counterpart. 

Perhaps autonomous racing drivers could be designed to consider the ‘personal injury’ (\textit{e.g.} magnitude of decelerations / impacts \textit{etc}.) they risk incurring - to provide a level playing field between man and machine. One could, \textit{e.g.} introduce a reinforcement learning system into the AV with penalties that scale with the riskiness of a potential crash,\textit{ e.g.} a high-energy, high-acceleration event could incur a large penalty versus a smaller one in a low-kinetic-energy risk manoeuvre. However, if performance becomes driven by algorithms, there may be concern that spectators find it more difficult to relate with the teams. 

How this plays out depends very much on how the role of AVs in motorsport is defined and regulated. The obvious model would be AVs vs. humans as hinted at by A2RL, pitting human instinct and risk versus algorithmic creativity and computational precision. The results shown in this work indicate that the technical capabilities of AVs are already approaching the point where if the balancing of risk-versus-reward (possibly using reinforcement learning) can be mastered, wheel-to-wheel human-versus-AV racing becomes a compelling, competitive prospect. Alternatively, a very different model would envisage a 'hybrid' approach, such as is seen in FSG, pairing up the human driver and autonomous driver so each team can benefit from both, having a first-hand comparison to help their development of each method. These formats could generate a new, fresh competition and narratives, whilst continuing to advance both AV technology and human driven skills.
\smallskip
\section{Conclusion}
\smallskip
This paper presented the question of whether autonomous vehicles could realistically compete ‘wheel-to-wheel’ with human drivers in a racing environment. While current evidence suggests that autonomous performance levels are very close to (and in some cases comparable with) human drivers in terms of reaction time and overall lap performance, important limitations still remain. For instance, autonomous racing vehicles do not generally model risk - and therefore are unable to make risk-aware decisions. The results of deploying autonomous vehicles on a track with human drivers would thus be unpredictable - with obvious safety concerns. If, in turn, this led to overly cautious regulations, it could potentially threaten the excitement of racing and take away from the traditional values of motorsport.

Encouraging data (A2RL on season-to-season improvement; FSG regarding skidpad; FSUK-AI on trajectories) shows AVs are hot on the tail of human drivers in terms of raw ‘against the clock’ performance, with significantly closer lap times over the last few years, and end-to-end latency now comparable with human reaction times. However, the limitations of AVs ‘higher level’ functions with respect to planning, racecraft and strategy - particularly when faced with challenging scenarios \textit{(e.g.} overtaking, avoiding incidents \textit{etc}.) - means wheel-to-wheel racing between (hu)man and machine remains a considerable future challenge. Notwithstanding, if the pace of development continues at the aggressive pace shown in the road-going AV market, perhaps AVs may be ready to compete against humans for the chequered flag in the next 5 years.

In order to realise the potential, future research directions will need to address several critical technological aspects - particularly with respect to safety, risk management, racecraft and strategy; with regulations developed accordingly. In future, with sufficient technological advancement and clearly-defined rule sets, it appears likely that AVs could compete alongside - and eventually surpass the capabilities of - human racing drivers.

\newpage

\bibliographystyle{apalike} 
\bibliography{references}

\end{document}